\documentclass[conference,a4paper]{IEEEtran}
\IEEEoverridecommandlockouts

\usepackage[hidelinks]{hyperref}
\usepackage[cmex10]{amsmath}
\usepackage{amssymb,amsfonts}
\usepackage{dblfloatfix}
\usepackage{enumitem}
\usepackage[ruled,vlined]{algorithm2e}
\usepackage{graphicx}
\graphicspath{{Figures/PDF/}{Figures/PNG/}}
\usepackage{tabularx}
\usepackage{booktabs}
\usepackage{siunitx}
\usepackage[numbers,compress]{natbib}
\usepackage{texnames}
\usepackage{bm,bbm}
\usepackage{orcidlink}
\usepackage{xcolor}

\begin{document}

\title{FlameVQA: A Physically-Grounded UAV Wildfire VQA Benchmark with Radiometric Thermal Supervision
  \thanks{This material is based upon work supported by the National Science Foundation under Grant Nos. CNS-2232048 and CNS-2204445, and by NASA under award No. 80NSSC23K1393.}
}

\author{
  \IEEEauthorblockN{Mobin Habibpour\orcidlink{0009-0001-9271-630X}}
  \IEEEauthorblockA{\textit{Clemson University}\\
    Clemson, SC, USA\\
  mhabibp@clemson.edu}
  \and
  \IEEEauthorblockN{John Spodnik}
  \IEEEauthorblockA{\textit{Clemson University}\\
    Clemson, SC, USA\\
  jspodni@g.clemson.edu}

    \and
  \IEEEauthorblockN{Niloufar Alipour Talemi\orcidlink{0009-0000-6881-3671}}
  \IEEEauthorblockA{\textit{Clemson University}\\
    Clemson, SC, USA\\
    nalipou@clemson.edu}

      \and
  \IEEEauthorblockN{Fatemeh Afghah\orcidlink{0000-0002-2315-1173}}
  \IEEEauthorblockA{\textit{Clemson University}\\
    Clemson, SC, USA\\
    fafghah@clemson.edu}
}

\maketitle

\begin{abstract}

Wildfire monitoring from UAVs requires reliable reasoning over complex aerial scenes, where smoke, scale variation, and occlusions often limit RGB-only interpretation. We introduce FlameVQA, a multiple-choice visual question answering benchmark for UAV-based wildfire intelligence built on FLAME 3, leveraging paired RGB imagery and radiometric thermal TIFFs for temperature-grounded, safety-critical reasoning. FlameVQA includes 34 multiple-choice questions per image spanning six operational capability groups, covering tasks such as detection, localization, distribution/coverage estimation, cross-modal reasoning, and flight planning. To ensure label reliability, we combine MLLM-assisted annotation with deterministic thermal rules and cross-question consistency checks, followed by human auditing. We also evaluate representative MLLMs on FlameVQA to provide baselines for future work. Results show strong performance when explicit cross-modal cues are available, but notable failures on presence detection under heavy smoke and on coverage estimation. These findings suggest that current MLLMs require domain-specific adaptation to better support disaster and wildfire monitoring. The dataset and benchmark code are open-source at \url{https://github.com/mobiiin/WildFire_VQA}

\end{abstract}

\begin{IEEEkeywords}
  Wildfire, UAV, Multimodal Large Language Models, Remote sensing VQA, Radiometric Thermal, Benchmark
\end{IEEEkeywords}

\section{Introduction}
\label{sec:intro}

Visual question answering (VQA) enables users to query visual data with natural language, requiring joint perception, grounding, and reasoning across vision and language \cite{antol2015vqa, marino2019ok, talemi2026agentic, boussaid2025visual}. In remote sensing, VQA supports flexible semantic access to aerial imagery for disaster monitoring, environmental assessment, and situational awareness. However, aerial VQA remains challenging due to extreme scale variation, cluttered scenes, viewpoint changes, and the need for spatially precise reasoning \cite{li2024hrvqa, alipour2025disa, talemi2025style}. These factors have motivated remote sensing benchmarks such as RSVQA \cite{zheng2021mutual} and HRVQA \cite{li2024hrvqa}, which emphasize high-resolution scenes and diverse question types.

\begin{figure}
    \centering
    \includegraphics[width=0.9\linewidth]{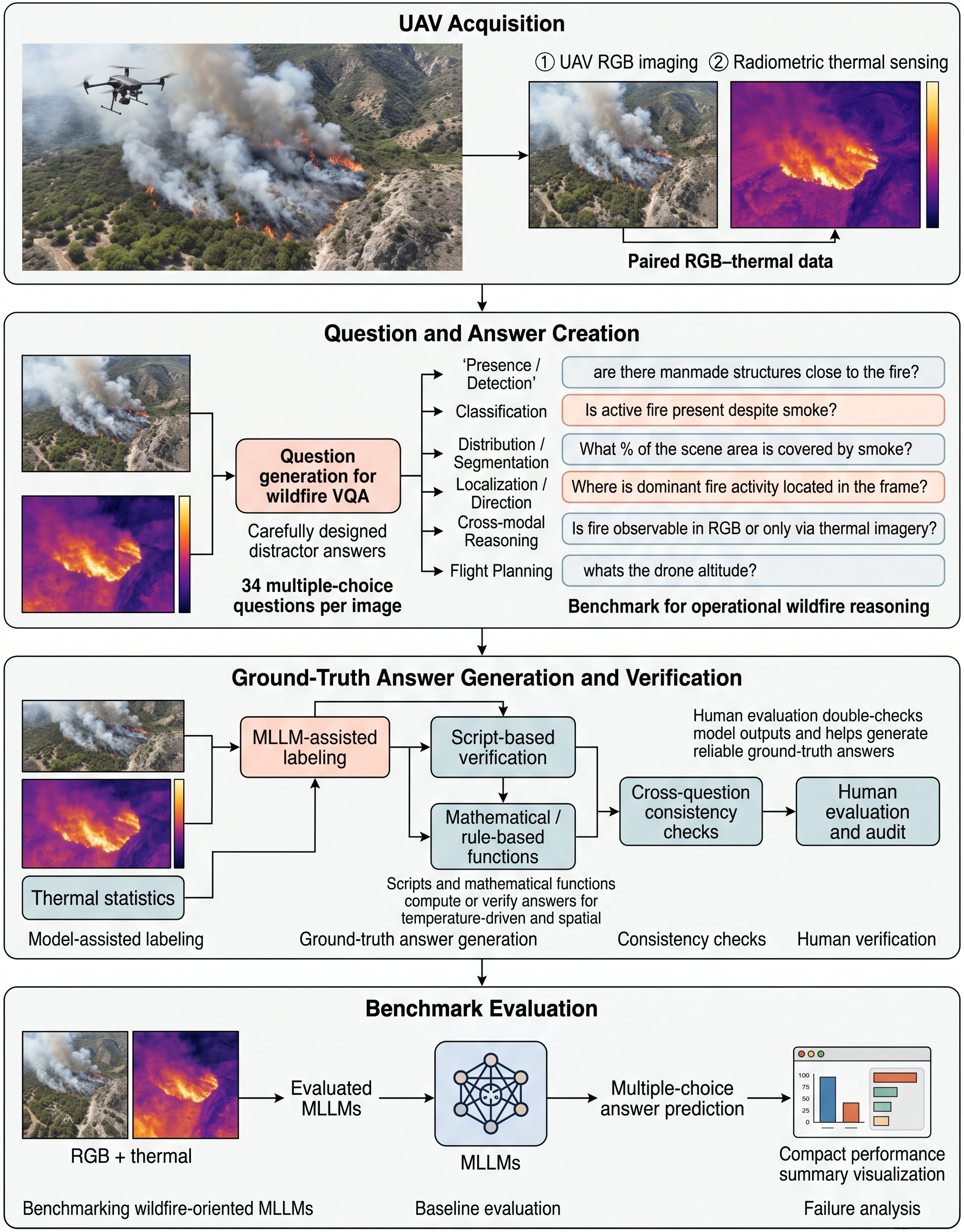}
\caption{\textbf{Overview of the FlameVQA Benchmark Pipeline.} (1) Paired RGB and radiometric thermal imagery of wildfires is captured via UAV. (2) Multiple-choice questions for operational reasoning across six capability groups are automatically generated. (3) Ground truth is produced using a hybrid engine integrating MLLM labeling, physics-based rules from thermal data, consistency checks, and human verification. (4) Wildfire-oriented MLLMs are evaluated with detailed failure analysis.}
    \label{fig:pipeline}
\end{figure}

Despite this progress, existing remote sensing VQA benchmarks largely focus on RGB optical imagery and general-purpose semantic queries, limiting evaluation in hazard-centric domains where modality-specific signals are critical. Wildfire monitoring is a prime example: analysts rely not only on visible flames and smoke patterns, but also on thermal signatures that reveal obscured hotspots, smoldering regions, and residual heat under canopy or smoke cover \cite{marinaccio2025seeing}. While UAV wildfire datasets include paired RGB and thermal imagery, they have mainly supported detection, segmentation, or classification, rather than multimodal question answering that integrates perception with higher-level reasoning.

In this work, we introduce \textbf{FlameVQA}, a visual question answering benchmark for UAV-based wildfire monitoring built on the FLAME~3 dataset \cite{Hopkins2024FLAME3}. FlameVQA leverages paired RGB imagery and radiometric thermal data collected during prescribed burns, enabling temperature-grounded reasoning in addition to visual interpretation. Each frame includes aligned RGB images, color-mapped thermal visualizations, and access to underlying radiometric thermal measurements. Figure~\ref{fig:pipeline} summarizes the end-to-end FlameVQA pipeline, from UAV acquisition to radiometric grounding, hybrid annotation, and benchmarking.

FlameVQA is a multiple-choice benchmark with 34 questions per image and carefully curated distractors, targeting scene understanding, hazard localization, coverage estimation, visibility assessment, and UAV-centric operational reasoning. We additionally benchmark representative MLLMs on FlameVQA, establishing a strong baseline for future comparisons and serving as a reference resource for the community. By grounding VQA in paired RGB and thermal UAV imagery and emphasizing operationally meaningful questions, FlameVQA complements existing aerial VQA benchmarks and facilitates the development of more robust, domain-aware VQA systems for disaster response. The main contributions of this paper can be summarized as follows:
\begin{itemize}[leftmargin=*]

\item We introduce FlameVQA, the first dedicated visual question answering benchmark for wildfire monitoring, built on paired RGB and radiometric thermal UAV imagery, enabling temperature-grounded and safety-critical reasoning beyond purely visual cues.

\item FlameVQA targets operational wildfire intelligence tasks, covering perception, localization, quantification, visibility assessment, and UAV-centric decision support, providing a realistic and challenging evaluation for multimodal models.

\item We evaluate representative multimodal large language models (MLLMs) on FlameVQA across multiple inference settings, revealing consistent limitations in safety understanding, multimodal grounding, and multi-image reasoning in state-of-the-art models.

\end{itemize}



\section{FlameVQA Benchmark Overview}
\label{sec:benchmark}

\subsection{Image Collection}
\begin{figure*}
    \centering
    \includegraphics[width=0.98\linewidth]{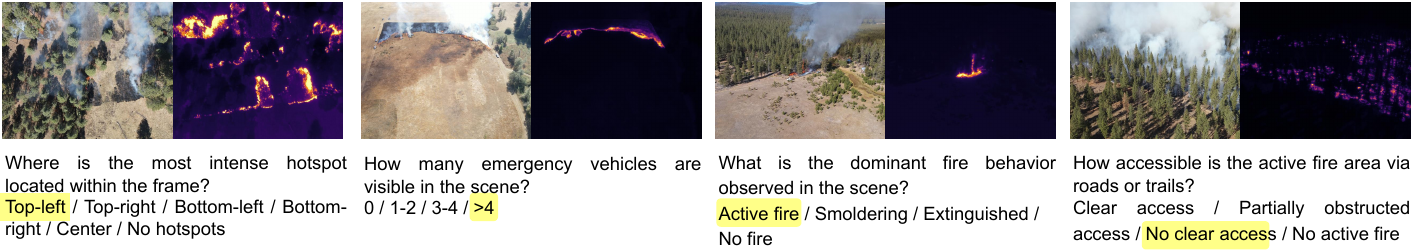}
    \caption{Random examples of image/question/answer triplets in FlameVQA.}
    \label{fig:examples}
\end{figure*}

FlameVQA is built on the public FLAME 3 computer vision subset, which provides synchronized aerial visible spectrum (RGB) imagery and radiometric thermal imagery collected by UAVs at rural prescribed burns \cite{Hopkins2024FLAME3}. A key advantage of FLAME 3 is that it provides radiometric thermal TIFFs with per-pixel temperature values, rather than palette-based thermal visualizations, enabling stronger validation for thermal-driven queries. In particular, hazardous hotspots can remain detectable in temperature even when flames are visually weak or partially obscured, such as when dense smoke masks visible combustion, while the TIFF still indicates concentrated regions exceeding high temperature thresholds. Radiometric thermal measurements are stored as single band thermal TIFF rasters, and a corresponding thermal JPEG is produced by mapping temperature values through a color map. We use paired RGB and IR images from three burns: Sycan Marsh, Willamette, and Shoetank. For each frame, we provide an RGB image, a color-mapped thermal visualization, and a radiometric thermal TIFF to support temperature-grounded queries, including obscured hotspots that are not clearly discernible in RGB imagery. Representative image--question--answer triplets from FlameVQA are shown in Figure~\ref{fig:examples}.

\subsection{Question Suite (Multiple-Choice)}
We design FlameVQA as a multiple-choice benchmark with carefully constructed distractor options, and we randomize option ordering to reduce language-model biases. This design follows recent MLLM evaluation practice showing that multiple-choice questions provide a standardized output format and are often more reliable for benchmarking than unconstrained free-form answers \cite{li2024seed, liu2024mmbench}. To assess whether MLLMs can serve as reliable assistants for UAV-based wildfire intelligence, it is necessary to evaluate a broad set of capabilities beyond fire presence detection. In operational settings, analysts must detect fire and smoke cues, interpret fire behavior, localize hazards, estimate impacted area, and reason about flight safety and observability limitations. 

Accordingly, we first prompt Gemini 2.5 Pro \cite{team2023gemini} with representative bimodal samples from FLAME 3 (RGB imagery and color-mapped thermal images) to generate an initial pool of about 50 candidate questions. We then iteratively curate these questions across diverse samples from the three burns, removing items that are unanswerable, rely on missing evidence, or provide limited operational value. Finally, we refine the answer options to ensure that distractors are informative and non-trivial while avoiding overly subjective choices, and we add expert-designed questions to address gaps in the Gemini-proposed pool. This process yields a benchmark of 34 multiple-choice questions per image spanning six wildfire-oriented capability groups.

\vspace{-0.3mm}
\begin{itemize}[leftmargin=*]
\item \textbf{Presence/Detection:} Determines whether wildfire cues and operationally relevant elements are present, covering fire/smoke, thermal activity, assets or infrastructure, and safety-relevant context.
\item \textbf{Classification:} Assigns discrete labels summarizing scene state, including fire behavior, vegetation, fuel condition, accessibility, and visible assets
\item \textbf{Distribution/Segmentation:} Characterizes the distribution of fire, fuels, vegetation, smoke, and heat, including pattern descriptors and coverage estimates above specified temperature thresholds.
\item \textbf{Localization/Direction:} Provides coarse spatial grounding and directional interpretation for hotspots, smoke origin, vegetation, and man-made elements, including cues from smoke drift.
\item \textbf{Cross-Modal Reasoning:} Evaluates joint reasoning over RGB and thermal signals, covering modality gaps, occlusion limits, and temperature-informed interpretation of fire activity.
\item \textbf{Flight Planning:} Captures UAV-centric reasoning, covering viewpoint/altitude, environmental constraints, safety risk near flames or smoke, and factors affecting flight.
\end{itemize}

Table~\ref{tab:dataset_summary} summarizes the dataset composition, modalities, and question suite used in this benchmark.

\begin{table}[t]
  \centering
  \caption{FlameVQA summary}
  \label{tab:dataset_summary}
  \footnotesize
  \begin{tabularx}{\columnwidth}{@{}l X@{}}
    \toprule
    \textbf{Property} & \textbf{Value} \\
    \midrule
    Source dataset & FLAME 3 CV subset \\
    Modalities per sample & RGB + Thermal JPEG + Radiometric thermal TIFF \\
    Burn sites & Sycan Marsh, Willamette Valley, Shoetank \\
    Total paired frames & $\sim$6,100 RGB--thermal pairs \\
    Total questions & 34 questions per image \\
    Question format & Multiple-choice \\
    Capability groups & Presence/Detection, Classification, Distribution/Segmentation, Localization/Direction, Cross-modal Reasoning, Flight Planning \\
    Label sources & MLLM + deterministic thermal rules + metadata \\
    \bottomrule
  \end{tabularx}
\end{table}

\section{Labeling and Quality Control}
\label{sec:labeling}

\subsection{Hybrid Labeling: MLLM + Deterministic Thermal Rules}

For dataset-wide labeling, we use Gemini 2.5 Pro \cite{team2023gemini} to generate initial answers for all questions across all multi-modal samples. Each labeling prompt includes the RGB image, the color-mapped thermal visualization, and a compact temperature summary computed from the radiometric thermal TIFF (e.g., minimum and maximum temperature, and threshold exceedance rates). The prompt then provides the question text and multiple-choice options, and instructs the model to output a single option identifier.

To ensure reliability across the full dataset, all generated labels are subsequently verified through a combination of deterministic, sensor-derived rules and human expert review. When available, deterministic physical signals are treated as the primary labeling reference, providing objective and reproducible supervision that avoids ambiguity in visual interpretation. For temperature-critical questions, such as threshold-based coverage estimation and fire behavior interpretation, labels are validated by directly reading the radiometric thermal TIFF using NumPy, enabling per-pixel temperature inspection and consistent verification of numeric and pattern-based answers. Similarly, UAV altitude above ground level is computed from EXIF GPS/altitude metadata combined with a terrain elevation model, yielding consistent altitude categories without reliance on subjective visual judgment.

\subsection{Physically Grounded Hotspot Definition (Canonical Example)}
\noindent Due to space constraints, we provide one canonical deterministic formulation used by multiple questions: thermal hotspot detection from radiometric TIFFs. Let $T(x,y)$ denote per-pixel temperature (in $^\circ$C). We form a binary hot mask via a fixed physical threshold:
\begin{equation}
  \mathcal{H}(x,y)=\mathbbm{1}\left[T(x,y)\ge \tau_{\mathrm{fire}}\right],
\end{equation}
and extract connected components as candidate hotspots. When altitude information is available, pixel areas are optionally converted to ground-projected areas using a field-of-view–based GSD approximation to suppress spurious detections and maintain consistency across UAV altitudes. This hotspot definition supports multiple question labels, including hotspot presence, extreme-temperature coverage bins, hotspot localization, and related spatial reasoning tasks.

\subsection{Cross-Question Consistency Checks}
\noindent To reduce label noise at scale, we apply simple within-frame logical constraints across semantically linked questions as part of dataset-wide verification. Examples include:
\vspace{-1mm}
\begin{itemize}[leftmargin=*]
  \item If \emph{No fire} is selected, hotspot-related questions must return \emph{No hotspots}.
  \item If \emph{No smoke} is selected, smoke coverage must be \emph{None}.
  \item If \emph{No structures} are detected, structure localization must return \emph{No structures visible}.
\end{itemize}
\vspace{-1mm}
Frames that violate these constraints are flagged for manual expert review and correction. This process is applied across the full dataset and provides a lightweight but effective quality control layer that complements deterministic rules and human verification. Algorithm~\ref{alg:labeling} provides an overview of the dataset-wide labeling and verification workflow.

\begin{algorithm}[t]
  \footnotesize
  \DontPrintSemicolon
  \caption{FlameVQA labeling and quality control (overview).}
  \label{alg:labeling}
  \KwIn{RGB+thermal image pair, radiometric TIFF, question set $\mathcal{Q}$}
  \KwOut{Final multiple-choice labels with QC flags}
  \BlankLine
  Compute thermal statistics from TIFF (thresholds, extrema)\;
  Generate initial labels using MLLM with RGB + IR + thermal stats\;
  \If{temperature-critical}{verify or override using deterministic TIFF rules\;}
  \If{metadata-grounded}{compute labels via EXIF and terrain data\;}
  Apply cross-question logical constraints\;
  Flag violations for human expert review and correction\;
  \Return{verified labels with QC flags}\;
\end{algorithm}

\section{Initial Validation Protocol and Baseline Results}
\label{sec:results}

\subsection{Targeted Human Evaluation on Challenging Questions}
\label{sec:human_eval}

\noindent Due to the cost of exhaustive manual annotation, we conduct a targeted human evaluation focusing on the most challenging FlameVQA questions: DS4, DS5, DS6 (distribution and coverage estimation), LD3 (spatial localization), CMR1 and CMR3 (cross-modal occlusion reasoning), and PD5 (fire presence under heavy smoke). These questions were selected because they require subjective judgment, aggregation over spatial regions, or cross-modal interpretation, and are therefore most likely to expose label noise or model bias. Table~\ref{tab:challenging_questions} summarizes the specific questions included in the targeted human evaluation.

Human evaluation was performed on the Willamette Valley subset, covering 12,733 evaluated question--answer pairs across the seven challenging questions listed in Table~\ref{tab:challenging_questions}. The Willamette Valley subset was chosen because it exhibits dense smoke, mixed fuel conditions, and frequent ambiguity between active fire and residual heat, making it a conservative stress test for both human and model interpretation. All human evaluations were performed by a graduate-level annotator with prior exposure to wildfire imagery, following the official FlameVQA question definitions and answer guidelines.

\begin{table}[t]
\centering
\caption{Subset of challenging FlameVQA questions analyzed in detail in this paper.}

\label{tab:challenging_questions}
\footnotesize 
\begin{tabularx}{\columnwidth}{@{}l X@{}} 
\toprule
\textbf{ID} & \textbf{Question Description} \\
\midrule
PD5  & Is active fire present despite heavy smoke or visual occlusion? \\
DS4  & What \% of scene area is covered by visible or thermal fire? \\
DS5  & What \% of the scene area is covered by smoke? \\
DS6  & What \% of the scene area contains active thermal hotspots? \\
LD3  & Where is dominant fire activity located in the frame? \\
CMR1 & Is fire observable in RGB or only via thermal imagery? \\
CMR3 & Is fire visibility consistent across RGB and thermal? \\
\bottomrule
\end{tabularx}
\end{table}

\subsection{VLM--Human Expert Comparison}
\label{sec:vlm_vs_human}

\noindent We compare the FlameVQA labels produced by the automated pipeline (MLLM-assisted labeling with rule-based verification) against a human expert on the Willamette evaluation subset for the seven challenging questions listed in Table~\ref{tab:challenging_questions}. Across all evaluated pairs, the automated labels match the human expert (treated as ground truth) in 70.78\% of cases (Table~\ref{tab:human_audit}). Agreement is substantially higher for cross-modal reasoning and spatial localization, while the lowest match rate occurs on PD5 (active fire under heavy smoke), which is inherently ambiguous even for experienced annotators and thus provides a conservative upper bound on achievable consistency. Many mismatches are near-boundary confusions (e.g., adjacent percentage bins or partial versus full occlusion), suggesting that most errors are semantically mild rather than catastrophic.

\begin{table}[t]
\centering
\caption{Automated label vs. human expert match rate on challenging questions (Willamette subset only).}

\label{tab:human_audit}
\resizebox{\columnwidth}{!}{%
\begin{tabular}{lccc}
\toprule
\textbf{Question ID} & \textbf{Category} & \textbf{Correct (\%)} & \textbf{Incorrect (\%)} \\
\midrule
PD5  & Presence / Detection        & 36.12 & 63.88 \\
DS5  & Distribution / Segmentation & 45.96 & 54.04 \\
LD3  & Localization / Direction    & 75.48 & 24.52 \\
DS4  & Distribution / Segmentation & 77.68 & 22.32 \\
CMR1 & Cross-modal Reasoning       & 82.68 & 17.32 \\
DS6  & Distribution / Segmentation & 86.53 & 13.47 \\
CMR3 & Cross-modal Reasoning       & 90.98 & 9.02 \\
\midrule
\textbf{Overall} & -- & \textbf{70.78} & \textbf{29.22} \\
\bottomrule
\end{tabular}}
\end{table}


\subsection{Open-Source Baseline Evaluation}
\label{sec:baseline}

\noindent We evaluate two representative open-source MLLMs as initial baselines on FlameVQA: LLaVA and Qwen-VL. These models cover two widely used design points for instruction-following vision-language reasoning, and they provide a practical reference for future comparisons on wildfire-oriented UAV imagery. Table~\ref{tab:opensource_vlm_bucketed} reports bucketed accuracy for LLaVA-1.6-7B and Qwen3-VL-8B-Instruct on the Willamette human-evaluated subset.

\begin{table}[t]
\footnotesize
\centering
\caption{Bucketed accuracy on the Willamette human-evaluated subset for two open-source MLLMs: LLaVA-1.6-7B and Qwen3-VL-8B-Instruct.}
\label{tab:opensource_vlm_bucketed}
\resizebox{\columnwidth}{!}{%
\begin{tabular}{lcc}
\toprule
\textbf{Subset} & \textbf{LLaVA Acc. (\%)} & \textbf{Qwen-VL Acc. (\%)} \\
\midrule
DS (DS4--DS6) & 18.07 & 25.18 \\
CMR (CMR1, CMR3) & 85.51 & 87.19 \\
PD+LD (PD5, LD3) & 14.18 & 45.33 \\
\midrule
\textbf{Overall (all 7)} & 36.23 & 48.65 \\
\bottomrule
\end{tabular}}
\end{table}

Qwen-VL consistently outperforms LLaVA across all evaluated buckets, with the largest gains observed in presence detection and spatial localization (PD+LD). Both models perform well on cross-modal reasoning questions, suggesting that joint RGB--thermal cues are effectively exploited when the task is explicit. In contrast, distribution and coverage estimation (DS) remains challenging for both models, highlighting the difficulty of spatial aggregation and percentage-based reasoning in wildfire scenes. Both models were evaluated using their publicly released checkpoints with fixed decoding parameters and identical input preprocessing across all questions.

\section{Conclusions}
\label{sec:conclusions}

In this paper, we introduced FlameVQA, the first UAV wildfire VQA benchmark that couples RGB imagery with radiometric thermal supervision to enable physically grounded, safety-critical reasoning. FlameVQA combines MLLM-assisted labeling with deterministic thermal rules and cross-question consistency checks, providing a scalable pipeline for benchmark construction under limited annotation budgets. FlameVQA establishes a foundation for rigorous evaluation of multimodal models in wildfire intelligence, and future work will expand site diversity, add expert-labeled subsets for high-stakes tasks, and release evaluation code and quality-control signals to support reproducible benchmarking.

\small
\bibliographystyle{IEEEtranN}
\bibliography{references}

@article{Hopkins2024FLAME3,
  title        = {FLAME 3 Dataset: Unleashing the Power of Radiometric Thermal UAV Imagery for Wildfire Management},
  author       = {Hopkins, Bryce and ONeill, Leo and Marinaccio, Michael and Rowell, Eric and Parsons, Russell and Flanary, Sarah and Nazim, Irtija and Seielstad, Carl and Afghah, Fatemeh},
  year         = {2024},
  journal      = {arXiv preprint arXiv:2412.02831},
  doi          = {10.48550/arXiv.2412.02831},
  url          = {https://arxiv.org/abs/2412.02831},
  archivePrefix= {arXiv},
  eprint       = {2412.02831},
  primaryClass = {cs.CV}
}

@inproceedings{antol2015vqa,
  title={Vqa: Visual question answering},
  author={Antol, Stanislaw and Agrawal, Aishwarya and Lu, Jiasen and Mitchell, Margaret and Batra, Dhruv and Zitnick, C Lawrence and Parikh, Devi},
  booktitle={Proceedings of the IEEE international conference on computer vision},
  pages={2425--2433},
  year={2015}
}

@inproceedings{marino2019ok,
  title={Ok-vqa: A visual question answering benchmark requiring external knowledge},
  author={Marino, Kenneth and Rastegari, Mohammad and Farhadi, Ali and Mottaghi, Roozbeh},
  booktitle={Proceedings of the IEEE/cvf conference on computer vision and pattern recognition},
  pages={3195--3204},
  year={2019}
}

@article{zheng2021mutual,
  title={Mutual attention inception network for remote sensing visual question answering},
  author={Zheng, Xiangtao and Wang, Binqiang and Du, Xingqian and Lu, Xiaoqiang},
  journal={IEEE Transactions on Geoscience and Remote Sensing},
  volume={60},
  pages={1--14},
  year={2021},
  publisher={IEEE}
}

@article{li2024hrvqa,
  title={HRVQA: A Visual Question Answering benchmark for high-resolution aerial images},
  author={Li, Kun and Vosselman, George and Yang, Michael Ying},
  journal={ISPRS Journal of Photogrammetry and Remote Sensing},
  volume={214},
  pages={65--81},
  year={2024},
  publisher={Elsevier}
}

@article{marinaccio2025seeing,
  title={Seeing Heat with Color--RGB-Only Wildfire Temperature Inference from SAM-Guided Multimodal Distillation using Radiometric Ground Truth},
  author={Marinaccio, Michael and Afghah, Fatemeh},
  journal={arXiv preprint arXiv:2505.01638},
  year={2025}
}

@inproceedings{li2024seed,
  title={Seed-bench: Benchmarking multimodal large language models},
  author={Li, Bohao and Ge, Yuying and Ge, Yixiao and Wang, Guangzhi and Wang, Rui and Zhang, Ruimao and Shan, Ying},
  booktitle={Proceedings of the IEEE/CVF Conference on Computer Vision and Pattern Recognition},
  pages={13299--13308},
  year={2024}
}

@inproceedings{liu2024mmbench,
  title={Mmbench: Is your multi-modal model an all-around player?},
  author={Liu, Yuan and Duan, Haodong and Zhang, Yuanhan and Li, Bo and Zhang, Songyang and Zhao, Wangbo and Yuan, Yike and Wang, Jiaqi and He, Conghui and Liu, Ziwei and others},
  booktitle={European conference on computer vision},
  pages={216--233},
  year={2024},
  organization={Springer}
}

@article{team2023gemini,
  title={Gemini: a family of highly capable multimodal models},
  author={Team, Gemini and Anil, Rohan and Borgeaud, Sebastian and Alayrac, Jean-Baptiste and Yu, Jiahui and Soricut, Radu and Schalkwyk, Johan and Dai, Andrew M and Hauth, Anja and Millican, Katie and others},
  journal={arXiv preprint arXiv:2312.11805},
  year={2023}
}

@inproceedings{talemi2025style,
  title={Style-Pro: Style-Guided Prompt Learning for Generalizable Vision-Language Models},
  author={Talemi, Niloufar Alipour and Kashiani, Hossein and Afghah, Fatemeh},
  booktitle={2025 IEEE/CVF Winter Conference on Applications of Computer Vision (WACV)},
  pages={6207--6216},
  year={2025},
  organization={IEEE}
}

@inproceedings{alipour2025disa,
  title={DiSa: Directional Saliency-Aware Prompt Learning for Generalizable Vision-Language Models},
  author={Alipour Talemi, Niloufar and Kashiani, Hossein and Nowdeh, Hossein R and Afghah, Fatemeh},
  booktitle={Proceedings of the 31st ACM SIGKDD Conference on Knowledge Discovery and Data Mining V. 2},
  pages={37--46},
  year={2025}
}

@article{talemi2026agentic,
  title={Agentic AI in Remote Sensing: Foundations, Taxonomy, and Emerging Systems},
  author={Talemi, Niloufar Alipour and Boone, Julia and Afghah, Fatemeh},
  journal={arXiv preprint arXiv:2601.01891},
  year={2026}
}

@inproceedings{boussaid2025visual,
  title={Visual question answering on multiple remote sensing image modalities},
  author={Boussaid, Hichem and Tosato, Lucrezia and Weissgerber, Flora and Kurtz, Camille and Wendling, Laurent and Lobry, Sylvain},
  booktitle={Proceedings of the IEEE/CVF Conference on Computer Vision and Pattern Recognition},
  pages={2344--2353},
  year={2025}
}

\end{document}